\documentclass[conference]{IEEEtran}
\IEEEoverridecommandlockouts
\usepackage{cite}
\usepackage{amsmath,amssymb,amsfonts}
\usepackage{algorithm}
\usepackage{algorithmic}
\usepackage{graphicx}
\usepackage{textcomp}
\usepackage{xcolor}
\usepackage{multirow}
\usepackage{csquotes}
\usepackage{rotating}

\newcommand\MyBox[2]{
  \fbox{\lower0.75cm
    \vbox to 1.7cm{\vfil
      \hbox to 1.7cm{\hfil\parbox{1.4cm}{#1\\#2}\hfil}
      \vfil}%
  }%
}

\newcommand{\boldx}{\bold{x}}
\newcommand{\Data}{\mathcal{D}}

\def\BibTeX{{\rm B\kern-.05em{\sc i\kern-.025em b}\kern-.08em
    T\kern-.1667em\lower.7ex\hbox{E}\kern-.125emX}}
\begin{document}

\title{Algorithm-Agnostic Interpretations for Clustering\\
}

\author{\IEEEauthorblockN{Christian A. Scholbeck\IEEEauthorrefmark{1}\IEEEauthorrefmark{3}}
\IEEEauthorblockA{
christian.scholbeck@stat.uni-muenchen.de}
\\
\IEEEauthorblockA{
\IEEEauthorblockA{\IEEEauthorrefmark{1}\textit{Department of Statistics} \\
\textit{Ludwig-Maximilians-Universität München}}}
\\
\IEEEauthorblockA{
\IEEEauthorblockA{\IEEEauthorrefmark{2}Munich Center for Machine Learning (MCML) \\
\textit{Ludwig-Maximilians-Universität München}}}
\\
\IEEEauthorblockA{\IEEEauthorrefmark{3} Authors contributed equally to this work.}%
\and
\IEEEauthorblockN{Henri Funk\IEEEauthorrefmark{1}\IEEEauthorrefmark{3}}
\IEEEauthorblockA{
henri.funk@stat.uni-muenchen.de}
\and
\IEEEauthorblockN{Giuseppe Casalicchio\IEEEauthorrefmark{1}\IEEEauthorrefmark{2}}
\IEEEauthorblockA{
giuseppe.casalicchio@lmu.de}
}
\maketitle

\begin{abstract}
A clustering outcome for high-dimensional data is typically interpreted via post-processing, involving dimension reduction and subsequent visualization. This destroys the meaning of the data and obfuscates interpretations. We propose algorithm-agnostic interpretation methods to explain clustering outcomes in reduced dimensions while preserving the integrity of the data.
The permutation feature importance for clustering represents a general framework based on shuffling feature values and measuring changes in cluster assignments through custom score functions. The individual conditional expectation for clustering indicates observation-wise changes in the cluster assignment due to changes in the data. The partial dependence for clustering evaluates average changes in cluster assignments for the entire feature space.
All methods can be used with any clustering algorithm able to reassign instances through soft or hard labels. In contrast to common post-processing methods such as principal component analysis, the introduced methods maintain the original structure of the features.
\end{abstract}

\begin{IEEEkeywords}
Interpretable clustering; algorithm-agnostic; permutation feature importance; individual conditional expectation; partial dependence; PFI; ICE; PD.
\end{IEEEkeywords}
\section{Introduction}

Recent efforts have focused on making machine learning models interpretable, both via model-agnostic interpretation methods and novel interpretable model types \cite{molnar_iml}, which is referred to as interpretable machine learning or explainable artifical intelligence in different contexts. Research in unsupervised learning - which includes clustering - has mostly avoided the topic of interpretability. However, it is desirable to explain why an observation was clustered in a certain way and what distinguishes clusters from each other \cite{plant_inconco}. Explanations increase human trust in decisions that are made on the basis of a clustering outcome and can be used to improve the underlying algorithm or its parameters. Unfortunately, the success in addressing the issue of cluster interpretability has been limited \cite{bertsimas_interpretable_clustering}. There are two options to receive interpretable clusters. Either using an algorithm that produces interpretable clusters \cite{plant_inconco, bertsimas_interpretable_clustering, lawless_multi_polytope_machine} or post-processing of the clustering results, which typically involves dimension reduction, e.g., via principal components analysis, and subsequent visualization of lower-dimensional data. The first option is restricted by the availability of clustering algorithms that produce interpretable clusters. The second option obfuscates interpretations by destroying the features used to cluster the data.

\subsection{Contributions}

This paper presents a novel, algorithm-agnostic way to interpret any clustering outcome while preserving the original features, inspired by model-agnostic interpretation techniques from supervised learning (SL).
\par
\textit{What do we mean by algorithm-agnostic interpretation?} 
We consider clustering interpretability to be any information that improves our knowledge of the clustering routine. This includes information regarding the relevance of features for the entire clustering outcome or for the constitution of single clusters. Our methods examine the current state of the clustering routine, conditional on a given data set. They are based on the principles of sensitivity analysis (SA) \cite{saltelli_sa}, changing feature values in a systematic way and evaluating changes in cluster assignments, thus being algorithm-agnostic for any clustering method that can assign instances to existing clusters through soft or hard labels. 
\par
The permutation feature importance (PFI) \cite{fisher_pfi, breiman_randomforests} serves as an inspiration for the PFI for clustering (PFIC), the individual conditional expectation (ICE) \cite{goldstein_ice} for the ICE for clustering (ICEC), and the partial dependence (PD) \cite{friedman_pd} for the PD for clustering (PDC). 

\subsection{Paper Outline}

We provide background information on model interpretations in SL and interpretable clustering in Section \ref{sec:background}. Section \ref{sec:notation} introduces the notation. In Sections \ref{sec:pfi}, \ref{sec:ice}, and \ref{sec:pd}, we define the PFIC, ICEC, and PDC, respectively. Section \ref{sec:notes} provides additional comments on the methodology behind our techniques. Section \ref{sec:application} demonstrates all methods on both simulated and real data. In Section \ref{sec:conclusion}, we discuss our results and provide an outlook on future work.

\section{Background and Related Work}
\label{sec:background}

With roots dating back several decades, the interpretation of model output has become a popular research topic only in recent years \cite{molnar_history}. Existing techniques provide interpretations or explanations (terms we use exchangeably in this paper) in terms of feature summary statistics or visualizations (e.g., a value indicating a feature's importance to the model or a curve indicating its effects on the prediction), model internals (e.g., beta coefficients for linear regression models), data points (e.g., counterfactual explanations \cite{wachter_counterfactuals}), or surrogate models (i.e., interpretable approximations to the original model) \cite{molnar_iml}. 
\par
Established methods to determine feature summaries comprise the ICE, PD, accumulated local effects (ALE) \cite{apley_ale}, local interpretable model-agnostic explanations (LIME) \cite{ribeiro_lime}, Shapley values \cite{strumbelj_shapley, lundberg_shap}, or the PFI. The functional analysis of variance (FANOVA) \cite{saltelli_sa, hooker_fanova} and Sobol indices \cite{sobol_index} of a high-dimensional model representation are powerful tools to quantify input influence on the model output in terms of variance but are limited by the requirement for independent inputs. Modifications to adapt the FANOVA for dependent inputs have had a limited success \cite{owen_shapley}. Shapley values - a concept from game theory - can address certain settings for dependent inputs, e.g., to identify non-influential inputs (factor fixing), or quantifying a feature's total order importance (including all types of interactions) \cite{iooss_shapley}.
\par
Unsupervised learning, including clustering, has largely been ignored by this line of research.
However, for high-dimensional data sets, the clustering routine can often be considered a black box, as we may not be able to assess and visualize the multidimensional cluster patterns found by the algorithm. It therefore is desirable to receive deeper explanations on how an algorithm makes its decisions and what differentiates clusters from each other.
Interpretable clustering algorithms incorporate the interpretability criterion directly into the cluster search. Interpretable clustering of numerical and categorical objects (INCONCO) \cite{plant_inconco} is an information-theoretic approach based on finding clusters that minimize minimum description length. It finds simple rule descriptions of the clusters by assuming a multivariate normal distribution and taking advantage of its properties. Interpretable clustering via optimal trees (ICOT) \cite{bertsimas_interpretable_clustering} uses decision trees to optimize a cluster quality measure. In \cite{lawless_multi_polytope_machine} clusters are explained by forming polytopes around them. Mixed integer optimization is used to jointly find clusters and define polytopes.
\par
Analogously to SL, we may define post-hoc interpretations as ones that are obtained after the clustering procedure, e.g., by showing a subset of representative elements of a cluster or via visualization techniques such as scatter plots or saliency maps \cite{kosara_interpretability}. Running another algorithm on top of the clustering outcome is referred to as post-processing. Typically, the data is high-dimensional and requires the use of dimensionality reduction techniques such as principal component analysis (PCA) before being visualized in two or three dimensions. PCA creates linear combinations of the original features called the principal components (PCs). The goal is to select fewer PCs than original features while still explaining most of their variance. PCA obscures the information contained in the original features by rotating the system of coordinates, thereby not revealing dependencies between features but instead between the PCs and the features. For instance, interpretable correlation clustering (ICC) \cite{achtert_correlation_clustering} uses post-processing of correlation clusters. A correlation cluster groups the data such that there is a common within-cluster hyperplane of arbitrary dimensionality. ICC applies PCA to each correlation cluster's covariance matrix, thereby revealing linear patterns inside the cluster.  One can also use an SL algorithm to post-process the clustering outcome which learns to find interpretable patterns between the found cluster labels and the features. Although we may use any SL algorithm, classification trees are a suitable choice due to naturally providing decision rules on how they arrive at a prediction \cite{bertsimas_interpretable_optimization}.

Another post-processing option is to conduct a form of SA where data are deliberately manipulated and reassigned to existing clusters. In \cite{ellis_g2pc}, the PFI is adapted for clustering, where feature values are first shuffled and the observations are reassigned to existing clusters. The percentage of change between clusters is used as a feature importance indication, termed global permutation percent change (G2PC) for global evaluations and local permutation percent change (L2PC) for evaluations of single instances. This methodology, originating in SA (where it is also referred to as the pick-freeze method due to picking select input values and freezing the remainders) and SL, is suited for algorithm-agnostic methods and provides the basis for the methods proposed in this paper.
\par
The PFIC represents a more general framework than G2PC which can be fine-tuned to the task at hand with a custom score metric and which is able to evaluate the cluster-specific feature importance. The ICEC provides a more targeted analysis than L2PC. Instead of shuffling in the same fashion as G2PC, the ICEC orders the feature values created to reevaluate cluster assignments, which can then be visualized for user-friendly interpretations. The PDC combines the motives behind the PFIC and ICEC. It globally explains effects on the clustering outcome for the entire feature space through aggregating local interpretations (the ICECs) and at the same time can be interpreted visually and more systematically than G2PC.

\section{Notation}

\label{sec:notation}
We cluster a data set $\Data = \left\{\boldx^{(i)}\right\}_{i = 1}^n$ where $\boldx^{(i)}$ denotes the $i$-th observation. A single observation $\boldx$ consists of $p$ feature values $\boldx = (x_1, \dots, x_p)$. A subset of features is denoted by $S \subseteq \{1, \dots, p\}$ with the complement set being denoted by $-S = \{1, \dots, p\} \;\setminus\; S$. Using $S$, an observation $\boldx$ can be partitioned so that $\boldx = (\boldx_S, \boldx_{-S})$. A data set $\Data$ where all features in $S$ have been shuffled is denoted by $\tilde{\Data}_S$. 
An algorithm generates $k$ clusters. The initial clustering is encoded within a function $f$ that - conditional on whether the clustering algorithm outputs hard or soft labels - maps each observation $\boldx$ to a cluster index $c$ (hard label) or to $k$ pseudo probabilities indicative of cluster membership  (soft labeling):

\begin{align*}
    \text{Hard labeling: } &f: \boldx \mapsto c, \; c \in \{1, \dots, k\} \\
    \text{Soft labeling: } &f: \boldx \mapsto [0, 1]^k
\end{align*}
For soft labeling algorithms, $f^{(c)}(\boldx)$ denotes the pseudo probability of observation $\boldx$ belonging to the $c$-th cluster. 
The same notation is used to indicate the cluster-specific value within an ICEC or PDC vector (see Sections \ref{sec:ice} and \ref{sec:pd}).

\section{Permutation Feature Importance \\ for Clustering}
\label{sec:pfi}

Shuffling a feature in the data set destroys the information it contains. The PFI is computed by evaluating the model performance before and after shuffling. The G2PC transfers this concept to clustering. It indicates the percentage of change between the cluster assignments of the original data and those from a permuted data set. A high G2PC indicates an important feature for the clustering outcome. However, for clusters that considerably differ in size, the G2PC does not accurately represent the importance of features, as it is dominated by the cluster with the most observations.
\par
We instead propose a more general framework, termed the permutation feature importance for clustering (PFIC).
The clustering task is viewed as a multi-class classification problem where the assignment to clusters after permuting feature values is evaluated using an appropriate score, e.g., the F1 score. 
Our framework consists of four stages: (1) running the clustering algorithm, (2) shuffling a subset of features (i.e., columns) $S$ in the data set $\Data$, (3) assigning the shuffled data $\tilde{\Data}_S$ to the clusters from step (1), and (4) measuring the change in cluster assignments through an appropriate score function $h(f(\Data), f(\tilde{\Data}_S))$. G2PC is a special case of our framework where the score corresponds to the global percentage of change. The PFIC for feature set $S$ corresponds to:
\begin{equation*}
\text{PFIC}_S = h(f(\Data), f(\tilde{\Data}_S))
\end{equation*}
In order to reduce variance in the estimate resulting from shuffling the data, one can shuffle $t$ times and evaluate the distribution, e.g., the median for a point estimate:
\begin{equation*}
\text{PFIC}_S = \text{median} \left(h(f(\Data), f(\tilde{\Data}^{(1)}_S)), \dots, h(f(\Data), f(\tilde{\Data}^{(t)}_S)) \right)
\end{equation*}

\begin{algorithm}
    \caption{Global PFIC}
    \label{alg:pfic_global}
    \begin{algorithmic}
    \STATE run clustering algorithm 
    \FORALL{$\text{iter} \in \{1, \dots, t\}$}
        \STATE shuffle columns $S$ 
        \STATE compute hard labels
        \STATE create multi-class confusion matrix
        \STATE compute score $h$ from confusion matrix
    \ENDFOR
    \STATE evaluate distribution of $\{h^{(\text{iter})}\}_{\text{iter} \in \{1, \dots, t\}}$
    \end{algorithmic}
\end{algorithm}

\begin{algorithm}
    \caption{Cluster-Specific PFIC}
    \label{alg:pfic_specific}
    \begin{algorithmic}
    \STATE run clustering algorithm 
   \FORALL{$\text{iter} \in \{1, \dots, t\}$}
        \STATE shuffle columns $S$ 
        \STATE compute hard labels
        \FORALL{$c \in \{1, \dots, k\}$}
            \STATE create binary confusion matrix
            \STATE compute score $h_c$ from confusion matrix
        \ENDFOR
    \ENDFOR
    \STATE evaluate distribution of $\{h_c^{(\text{iter})}\}_{\text{iter} \in \{1, \dots, t\}}$
    \end{algorithmic}
\end{algorithm}

\subsection{Multi-Class Classification Problem}

When comparing original cluster assignments and the ones after shuffling the data, we can create a confusion matrix in the same way as in multi-class classification (see Table \ref{tab:multi_confusion_matrix}).  Multi-class classification performance can be evaluated by aggregating binary class comparisons - class $c$ versus the remaining classes - through a micro and macro score (see Appendix). The micro score is a suitable metric if all instances shall be considered equally important. The macro score suits a setting where all classes (i.e., clusters in our case) shall be considered equally important. The micro F1 score (see Appendix) is equivalent to classification accuracy (for settings where each instance is assigned a single label), so the following relation holds:
$$
\text{F1}_{\text{micro}} = 1 - \text{G2PC}
$$
Instead of aggregating binary comparisons, we can also directly evaluate them for cluster-specific interpretation purposes (see Table \ref{tab:binary_confusion_matrix}). Analogously, one can use established binary score metrics, e.g., the F1 score, Rand \cite{rand_index} or Jaccard \cite{jaccard_index} index. This allows much more flexible interpretations than G2PC. Algorithm \ref{alg:pfic_global} describes the global PFIC algorithm. Algorithm \ref{alg:pfic_specific} describes the cluster-specific PFIC algorithm.

\begin{table}[h]
    \centering
     \caption{Multi-class confusion matrix to compute the PFIC.}
    \label{tab:multi_confusion_matrix}
    \noindent
    \renewcommand\arraystretch{1.5}
    \setlength\tabcolsep{0pt}
    \begin{tabular}{p{0.025\textwidth} p{0.075\textwidth} p{0.15\textwidth} p{0.1\textwidth} p{0.1\textwidth}}
        \multirow{10}{*}{\rotatebox{90}{\parbox{5cm}{\bfseries\centering Cluster after shuffling}}} & 
        & \multicolumn{2}{c}{\bfseries Cluster before shuffling} & \\
        & & \bfseries Cluster 1 & $\dots$ & \bfseries Cluster c\\
        &  \bfseries Cluster 1 & \MyBox{$\#_{11}$}{} & $\ldots$ & \MyBox{$\#_{1c}$}{} \\[2.4em]
        &  $\vdots$ & $\vdots$ & $\ddots$ & $\vdots$ \\[1em]
        & \bfseries Cluster c & \MyBox{$\#_{c1}$}{} & $\ldots$ &  \MyBox{$\#_{cc}$}{}
    \end{tabular}
\end{table}

\begin{table}[h]
    \centering
        \caption{Binary confusion matrix to compute the PFIC.}
    \label{tab:binary_confusion_matrix}
    \noindent
    \renewcommand\arraystretch{1.5}
    \setlength\tabcolsep{0pt}
    \begin{tabular}{p{0.025\textwidth} p{0.1\textwidth} c p{0.1\textwidth} p{0.1\textwidth}} 
        \multirow{10}{*}{\rotatebox{90}{\parbox{5cm}{\bfseries\centering Cluster after shuffling}}} & 
        & \multicolumn{2}{c}{\bfseries Cluster before shuffling} & \\
        & & \bfseries Cluster c & \bfseries $\overline{\text{Custer c}}$  & \\
        &  \bfseries Cluster c & \MyBox{$\#_{cc}$}{} & \MyBox{$\#_{c\overline{c}}$}{} & \\[1.9em]
        & \bfseries $\overline{\text{Cluster c}}$ & \MyBox{$\#_{\overline{c}c}$}{} & \MyBox{$\#_{\overline{cc}}$}{} &
    \end{tabular}
\end{table}

\par \textit{How to interpret the PFIC:} The global PFIC indicates how shuffling the values of a feature or a set of features changes the assignment of instances to the clusters, measured by a multi-class score metric. The multi-class score is a global feature importance value that ranks features according to their overall contribution to the clustering outcome.
The cluster-specific PFIC indicates how shuffling the values of a feature or a set of features changes the assignment of instances to that specific cluster versus the remaining ones, measured by a binary score metric. The binary score is a regional feature importance value that ranks the feature contributions to observations being assigned to a specific cluster.

\section{Individual Conditional Expectation \\ for Clustering}
\label{sec:ice}
The ICE indicates the prediction of an SL model for a single observation $\boldx$ where a subset of values $\boldx_S$ is replaced with values $\tilde{\boldx}_S$ while we condition on the remaining features $\boldx_{-S}$, i.e., keep them fixed.
We propose the ICEC which works for both soft labeling and hard labeling clustering algorithms.
For soft labeling (sICEC), it corresponds to the pseudo probability that an observation $\boldx$ with replaced values $\tilde{\boldx}_S$ is assigned to the $k$-th cluster. For hard labels (hICEC), it indicates the cluster assignment of an observation $\boldx$ with replaced values $\tilde{\boldx}_S$:
\begin{align*}
    \text{ICEC}_\boldx(\tilde{\boldx}_S) &= f(\tilde{\boldx}_S, \boldx_{-S})
\end{align*}
For soft labeling algorithms, the sICEC corresponds to a k-way vector:
$$
\text{sICEC}_\boldx(\tilde{\boldx}_S) = \left(f^{(1)}(\tilde{\boldx}_S, \boldx_{-S}), \dots, f^{(k)}(\tilde{\boldx}_S, \boldx_{-S})\right)
$$
The sICEC is best interpreted visually (see Section \ref{sec:application}). Depending on the initial cluster assignment, sICECs will have a similar shape (see Fig. \ref{fig:cpdcice}). Different shapes indicate interactions with other features.

\begin{algorithm}
    \caption{ICEC}
    \label{alg:icec}
    \begin{algorithmic}
    \STATE run clustering algorithm
    \STATE sample $m$ vectors of feature values $\{\tilde{\boldx}^{(j)}_S\}_{j \in \{1, \dots, m\}}$
    \FORALL{$i \in \{1, \dots, n\}$} 
        \FORALL{$j \in \{1, \dots, m\}$}
            \STATE generate hypothetical observation $\boldx$ = $(\tilde{\boldx}^{(j)}_S, \boldx_{-S}^{(i)})$  
            \STATE $\text{ICEC}_{\boldx^{(i)}}(\tilde{\boldx}^{(j)}_S) = f(\boldx)$ 
        \ENDFOR
    \ENDFOR
    \end{algorithmic}
\end{algorithm}

\par \textit{How to interpret the ICEC:} The sICEC indicates how replacing values of a feature or a set of features changes the pseudo probability of a single instance being assigned to each existing cluster. The hICEC indicates how replacing values of a feature or a set of features changes the hard cluster assignment of a single instance.

\section{Partial Dependence for Clustering}
\label{sec:pd}

\noindent The partial dependence (PD) \cite{friedman_pd} represents the expectation of an SL model w.r.t. a subset of features. The PD can be estimated through a point-wise aggregation of ICEs. We propose the PDC which indicates both cluster-specific and global effects of a subset of features on the clustering. Analogously to the ICEC, it works for both soft labeling (sPDC) and hard labeling (hPDC):
\begin{align*}
    \text{hPDC}_\Data(\tilde{\boldx}_S) &= \text{mode} \left(\{ \text{ICEC}_{\boldx^{(1)}}(\tilde{\boldx}_S), \dots, \text{ICEC}_{\boldx^{(n)}}(\tilde{\boldx}_S) \}\right) \\
    \text{sPDC}_\Data(\tilde{\boldx}_S) &= \left( \overline{\text{sICEC}}^{(1)}_{\Data}(\tilde{\boldx}_S) , \dots, \overline{\text{sICEC}}^{(k)}_{\Data}(\tilde{\boldx}_S) \right)
\end{align*}
where
$$
\overline{\text{sICEC}}^{(c)}_{\Data}(\tilde{\boldx}_S) = \frac{1}{n} \sum_{i = 1}^n \text{sICEC}^{(c)}_{\boldx^{(i)}}(\tilde{\boldx}_S)
$$
Alternatively, one can use the median instead of the mean value. Due to changing feature values for all observations and then aggregating the ICECs, the PDC is a global explanation for the entire feature space.
The PDC is best interpreted visually (see Section \ref{sec:pd}). For a hard labeling algorithm, the outcome of the ICEC and thus also PDC is a scalar indicating cluster membership. We receive $n$ hard labels for each replaced feature value. A useful interpretation (and visualization) is to evaluate the percentage of majority vote labels, indicating \enquote{certainty} of the PDC for hard labeling algorithms (see Fig.  \ref{fig:hpdc}). The percentage of majority voted class labels indicates the homogeneity of ICECs. If substituting a feature set by the same values for all observations results in a reassignment to one cluster for the majority of instances, the PDC is a good interpretation instrument. Otherwise, further investigations into the ICECs are required.

\begin{algorithm}
    \caption{sPDC}
    \label{alg:pdc}
    \begin{algorithmic}
    \STATE run clustering algorithm
     \STATE sample $m$ vectors of feature values $\{\tilde{\boldx}^{(j)}_S\}_{j \in \{1, \dots, m\}}$\FORALL{$i \in \{1, \dots, n\}$}
        \STATE compute $\text{ICEC}_{\boldx^{(i)}}$ (see Algorithm \ref{alg:icec})
    \ENDFOR
    \FOR{$j \in \{1, \dots, m\}$}
        \FOR{$c \in \{1, \dots, k\}$}
            \STATE $\text{sPDC}^{(c)}_{\Data}(\tilde{\boldx}^{(j)}_S) = \frac{1}{n} \sum_{i = 1}^n \text{sICEC}^{(c)}_{\boldx^{(i)}}(\tilde{\boldx}^{(j)}_S)$
        \ENDFOR
    \ENDFOR
    \end{algorithmic}
\end{algorithm}

\begin{algorithm}
    \caption{hPDC}
    \label{alg:pdc}
    \begin{algorithmic}
  \STATE run clustering algorithm
     \STATE sample $m$ vectors of feature values $\{\tilde{\boldx}^{(j)}_S\}_{j \in \{1, \dots, m\}}$\FORALL{$i \in \{1, \dots, n\}$}
        \STATE compute $\text{ICEC}_{\boldx^{(i)}}$ (see Algorithm \ref{alg:icec})
    \ENDFOR
    \FOR{$j \in \{1, \dots, m\}$}
        \STATE $\text{hPDC}_{\Data}(\tilde{\boldx}^{(j)}_S) = \text{mode}\left(\{\text{sICEC}_{\boldx^{(i)}}(\tilde{\boldx}^{(j)}_S)\}_{i \in \{1, \dots, n\}}\right)$
    \ENDFOR
    \end{algorithmic}
\end{algorithm}

\par \textit{How to interpret the PDC:} The sPDC indicates how - on average - replacing values of a feature or a set of features changes the pseudo probability of an observation being assigned to each existing cluster. The hPDC indicates how - on average - replacing values of a feature or a set of features changes the hard cluster assignment of an observation.

\section{Additional Notes on the Methodology}
\label{sec:notes}

\subsection{Sampling}

We sample values for a feature set $S$. A simple option is to use the sampling distribution, i.e., all observed values. In SA, one typically intends to explore the feature space as thoroughly as possible (space-filling designs). In SL, there are valid arguments against space-filling designs due to resulting in model extrapolations, i.e., predictions in areas where the model was not trained with enough data \cite{hooker_permutation, molnar_pitfalls}. In clustering, the absence of a model and in turn the absence of associated model performance issues allow us to fill the feature space as extensively as possible, e.g., with unit distributions, random, or quasi-random (also referred to as low-discrepancy) sequences (e.g., Sobol sequences) \cite{saltelli_sa}. In fact, assigning unseen data to the clusters serves our purpose of visualizing the decision boundaries between them.
\par
\textit{Sampling strategy for each method:} For the PFIC, we evaluate a fixed dataset and simply shuffle $\boldx_S$. For the ICEC and PDC, we can either use observed values or strive for a more space-filling design. The more values we sample, the better our interpretations but the higher the computational cost.

\subsection{Reassigning versus Reclustering}

\begin{figure}[h]
    \centering
    \includegraphics[width=0.5\textwidth]{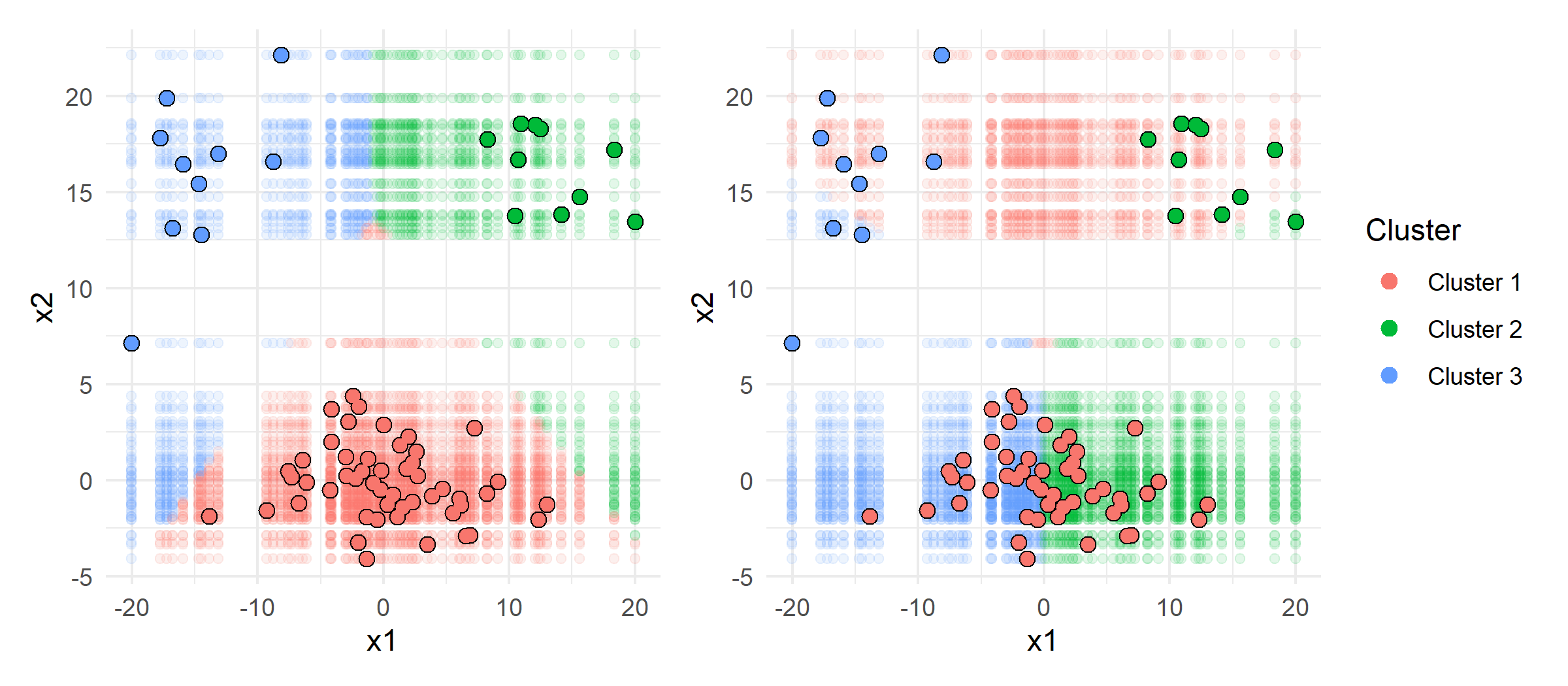}
    \caption{Observations (solid points) and Cartesian product (transparent grid) reassigned (left plot) and reclustered (right plot).}\label{fig:reclvsref}
\end{figure}

As discussed in \cite{ellis_g2pc}, one could argue that assigning instances to existing clusters violates the purpose behind clustering algorithms (which would form new clusters instead). Reclustering data that were manipulated on a greater scale results in a \enquote{concept drift} and different clusters. In Fig. \ref{fig:reclvsref} (left), we evaluate the Cartesian product of bivariate data that forms 3 clusters (grid lines).  The right plot visualizes a reclustering of the same Cartesian product, resulting in clearly visible changes in the shape and scale of the clusters.
\par
The PFIC, ICEC, and PDC can also be used to evaluate a reclustering of the data. For instance, one could compare the sets of observations clustered together before and after the change in the data (in all three methods). However, such an analysis is computationally considerably more costly. Furthermore, one must factor out stochasticity when reclustering, e.g., due to randomly initializing the initial cluster parameters.
\par
However, the influence of small and isolated changes in the data on the constitution of the clusters is negligible. For instance, the ICEC evaluates changes of single instances one-at-a-time, and the changes are typically restricted to single features or small sets of features. Although the PFIC and PDC evaluate simultaneous changes in all data instances, these are also restricted to a (typically very small) subset of features. Such a scenario is markedly different from increasing the size of the data as in Fig. \ref{fig:reclvsref}. However, note that - especially for simultaneous changes in the data - reassigning instances is not necessarily predictive of how a reclustering would appear. 
It follows that our methods are mainly targeted at characterizing high-dimensional clusters in reduced dimensions, conditional on a given dataset. In other words, we treat the found clustering outcome as a model, and our interpretations are akin to model-agnostic interpretations in SL.

\subsection{Algorithm-Agnostic Interpretations}

How to reassign instances differs across clustering algorithms. For instance, in $k$-means we assign an instance to the cluster with the lowest Euclidean distance; in probabilistic clustering such as Gaussian mixture models we select the cluster associated with the largest probability; in hierarchical clustering we select the cluster with the lowest linkage value, etc. \cite{ellis_g2pc}. In other words, although the implementation of the reassignment stage in our methods differs across algorithms (the computation of soft or hard labels), our interpretation techniques stay exactly the same. For our methods to be truly algorithm-agnostic, we incorporate variants to accommodate hard labeling algorithms as well.

\section{Application}
\label{sec:application}

\subsection{Simulations}

\subsubsection{Micro F1 versus Macro F1}

\begin{figure}[h]
    \centering
    \includegraphics[width=0.5\textwidth]{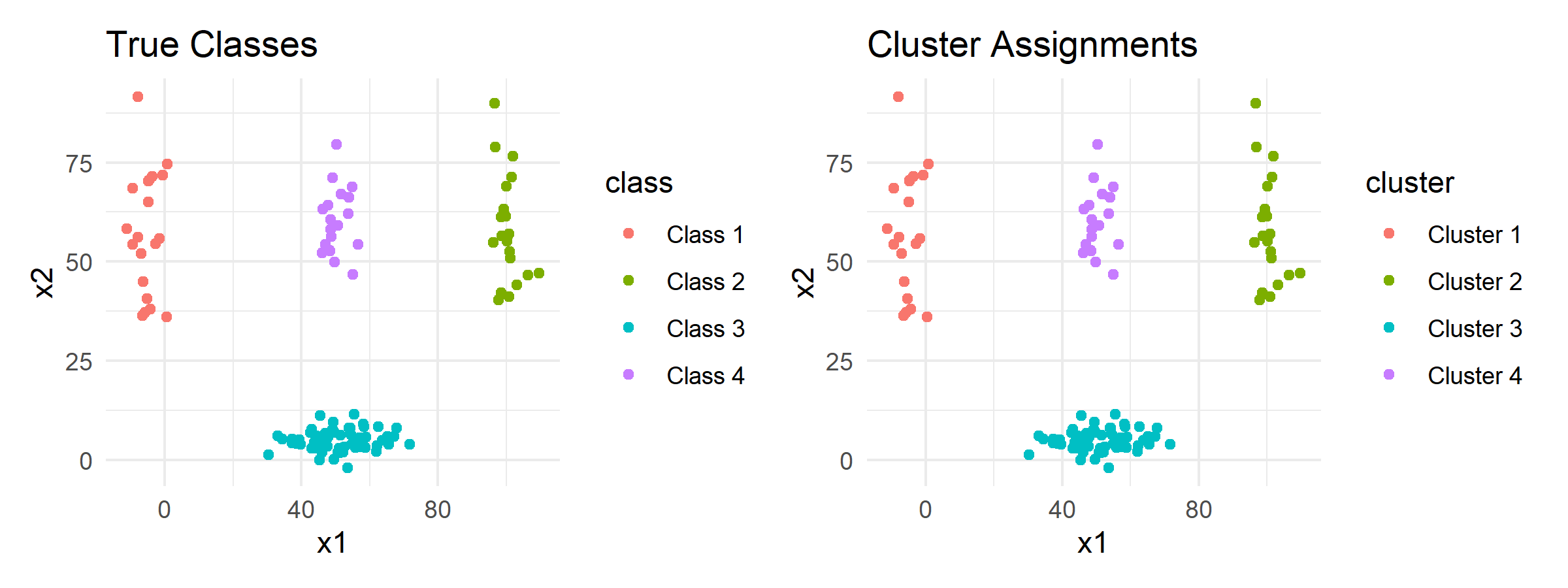}
    \caption{Visualization of the data and the perfect clustering of $c$-means.}\label{fig:imbcl}
\end{figure}

We simulate an imbalanced data set consisting of 4 classes (see Fig.  \ref{fig:imbcl}), where each class follows a different bivariate normal distribution. 60 instances are sampled from class 3 while 20 instances are sampled from each of the remaining classes. To capture the latent class variable, c-means is set to the 4 centers. The right plot in Fig.  \ref{fig:imbcl} displays the perfect cluster assignments. We can see that $x_1$ is the defining feature of the clustering for 3 out of 4 clusters.

We now compare the macro F1 and micro F1 score for $x_1$ and $x_2$. In Table \ref{tab:mivma}, we can see that micro F1 indicates that both features are equally important. Note that the micro F1 score corresponds to 1 - G2PC, which implies that G2PC is unable to identify the feature importance in this case. Macro F1 on the other hand is different for both features, indicating that $x_1$ is more important. Note that the F1 score is a similarity index. A low F1 score indicates a high dissimilarity between original data and shuffled data and thus a high feature importance.

\begin{table}[h]
\centering
  \caption{Global micro and macro F1 score for $x_1$ and $x_2$.}
    \label{tab:mivma}
\begin{tabular}{lrrr}
 \hline
 \multicolumn{4}{l}{macro F1}\\
 \hline
  features & 5\% quantile & median &95\% quantile\\  
 \hline
 $x_1$ & 0.36 & 0.43 & 0.49 \\ 
 $x_2$ & 0.58 & 0.64 & 0.70 \\ 
 \hline
 \multicolumn{4}{l}{micro F1 (accuracy)}\\
 \hline
  features & 5\% quantile & median &95\% quantile\\  
 \hline
 $x_1$ & 0.53 & 0.58 & 0.62 \\ 
 $x_2$ & 0.53 & 0.58 & 0.66 \\ 
 \hline
 \end{tabular}

\end{table}
These results stem from the fact that micro F1 accounts for each instance with equal importance. Cluster 3 is over-represented with 3 times as many observations as the remaining clusters. The macro F1 score accurately captures this by treating each cluster equally important, regardless of its size.

\subsubsection{Global versus Cluster-Specific PFIC}

\begin{figure}[h]
    \centering
    \includegraphics[width=0.5\textwidth]{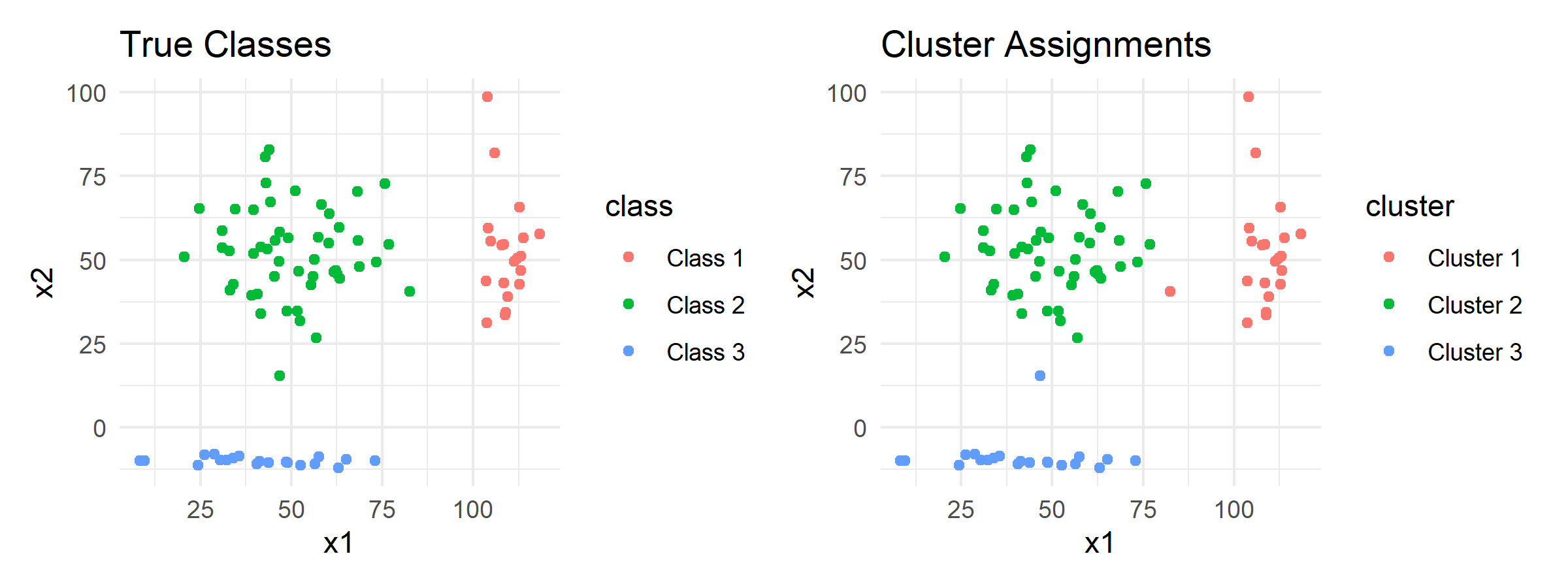}
    \caption{Three classes with different distributions clustered by c-means. True classes (left) and clusters (right) almost perfectly match.}\label{fig:cshap}
\end{figure}

We simulate three visibly distinctive classes (left plot in Fig.  \ref{fig:cshap}) where each follows a bivariate normal distribution with different mean and covariance matrices. 50 instances are sampled from class 2 and 20 instances are sampled from class 1 and class 3 each. We initialize c-means at the 3 mean values.
As shown in Fig.  \ref{fig:cshap}, the cluster assignments capture all three classes almost perfectly, except for an instance of class 2 being assigned to cluster 1 and one to cluster 3.
\par

We compare the global macro F1 (which weights the importance of clusters equally) to the cluster-specific F1 score (using the binary classification representation from Table \ref{tab:binary_confusion_matrix}). Table \ref{tab:mf1cfi} displays the global PFIC using the global macro F1 score. With a  median score of 0.62 for $x_1$ and 0.66 for $x_2$ in addition to the overlapping quantiles, there is no difference between the importance of both features for the clustering outcome.

\begin{table}[h]
\centering
    \caption{Global macro F1 feature importance for $x_1$ and $x_2$}
    \label{tab:mf1cfi}
\begin{tabular}{l rrr}
  \hline
 features & 5\% quantile & median &95\% quantile\\ 
  \hline
 $x_1$ &  0.55 & 0.62 & 0.69 \\ 
 $x_2$ & 0.59 & 0.66 & 0.75  \\ 
   \hline
\end{tabular}

\end{table}

In contrast, the cluster-specific PFIC offers a more detailed view of the contributions of each feature to the clustering outcome (see Table \ref{tab:ccfi}). 
While $x_1$ and $x_2$ are equally important in forming cluster 2, feature $x_2$ is considerably more important for cluster 3, and $x_1$ is the defining feature of cluster 1. Note that the mean scores per feature strongly resemble the macro scores from Table \ref{tab:mf1cfi}.

\begin{table}[h]
\centering
\caption{Regional class-specific feature importance for $x_1$ and $x_2$}
    \label{tab:ccfi}

\begin{tabular}{l rrr | r}
  \hline
 features & cluster 1 & cluster 2 & cluster 3 & mean similarity\\ 
  \hline
 $x_1$ & 0.24 & 0.73 & 0.86 & 0.61 \\ 
 $x_2$ &  1.00 & 0.73 & 0.26 & 0.66 \\ 
   \hline
\end{tabular}
\end{table}

\begin{figure}[h]
    \centering
    \includegraphics[width=0.5\textwidth, trim = {0 0 0 1cm}, clip]{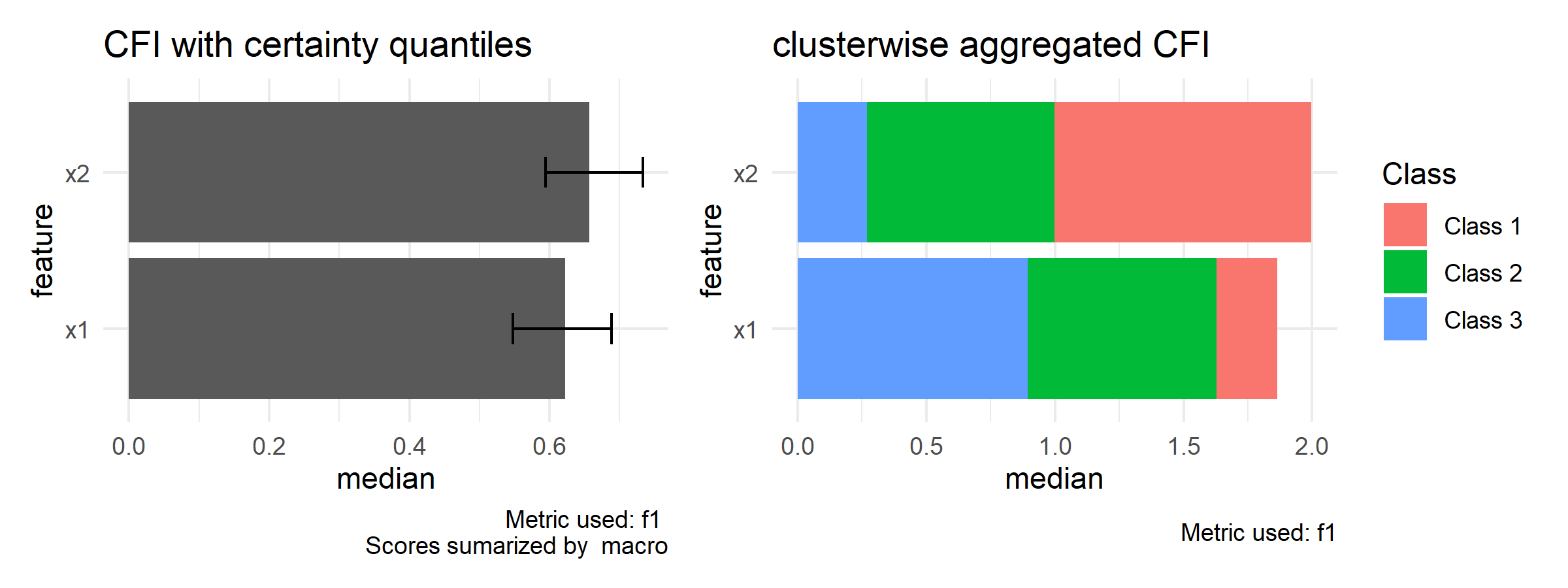}
    \caption{Global importance (left) compared to stacked cluster-specific importance (right).}\label{fig:cimpp}
\end{figure}

In Fig.  \ref{fig:cimpp}, we stack the cluster-specific PFIC for each feature, which results in similar total proportions between $x_1$ and $x_2$. The left bar plot shows the results for the macro F1 score. The whiskers represent the interquartile range.
The left plot indicates that both features are of equal global importance to the clustering, while the right bar plot reveals differences in cluster-specific importance of both features.

\subsubsection{ICEC and PDC}
\begin{figure}[h]
    \centering
    \includegraphics[width=0.5\textwidth]{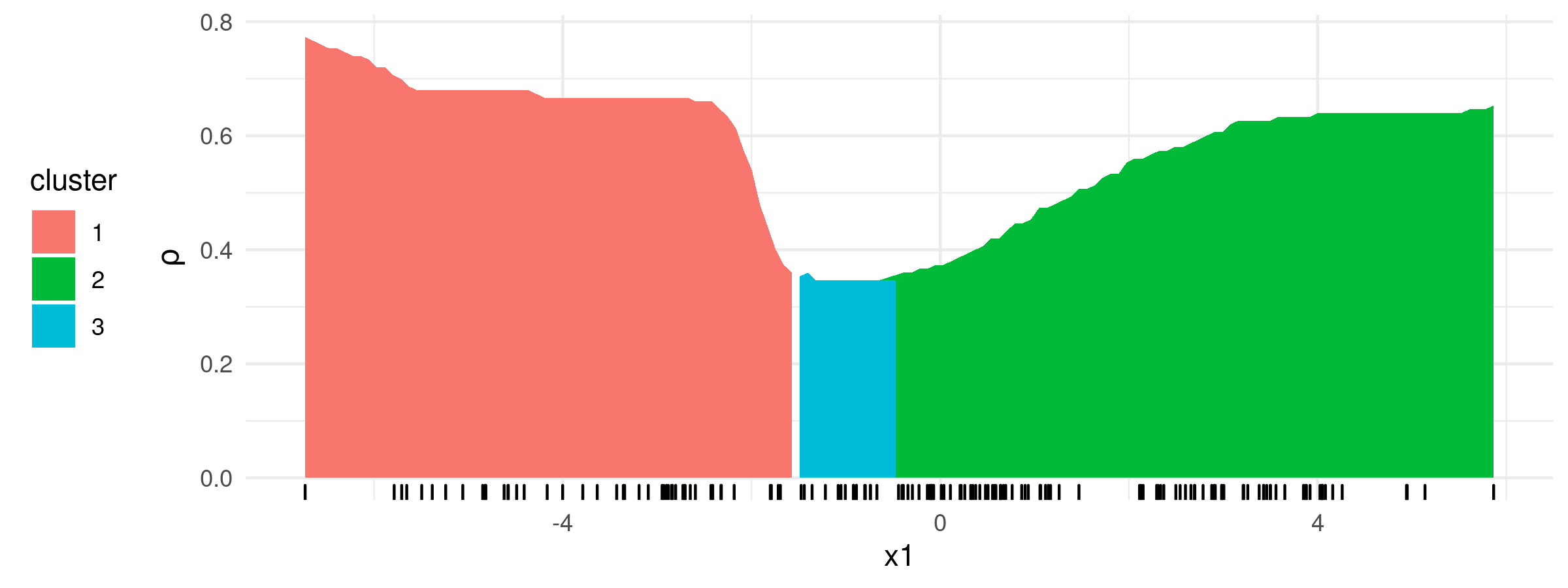}
    \caption{A plot indicating \enquote{certainty} of the hPDC. On average, replacing $x_1$ by the axis value results in an observation being assigned to the color-indicated cluster. The vertical distance indicates how many observations are assigned to the majority cluster.    \label{fig:hpdc}}
\end{figure}

We now draw 50 instances from three multivariate normally distributed classes.
To make them differentiable for the clustering algorithm, the classes are generated with an antagonistic mean structure. The covariance matrix of the three classes is sampled using a Wishart distribution (see Appendix for details).
The left plot in Fig.  \ref{fig:pdpsim} depicts the 3-dimensional distribution of the classes. We intend class 3 to be dense and classes 1 and 2 to be less dense but large in hypervolume. We initialize c-means at the 3 centers and optimize via Euclidean distance. Fig.  \ref{fig:pdpsim} visualizes the perfect clustering. Fig.  \ref{fig:hpdc} displays an hPDC plot for $x_1$ (see Section \ref{sec:pd}), indicating the majority vote of observations when exchanging values of $x_1$ on average for all observations.

\begin{figure}[h]
    \centering
    \includegraphics[width=0.5\textwidth]{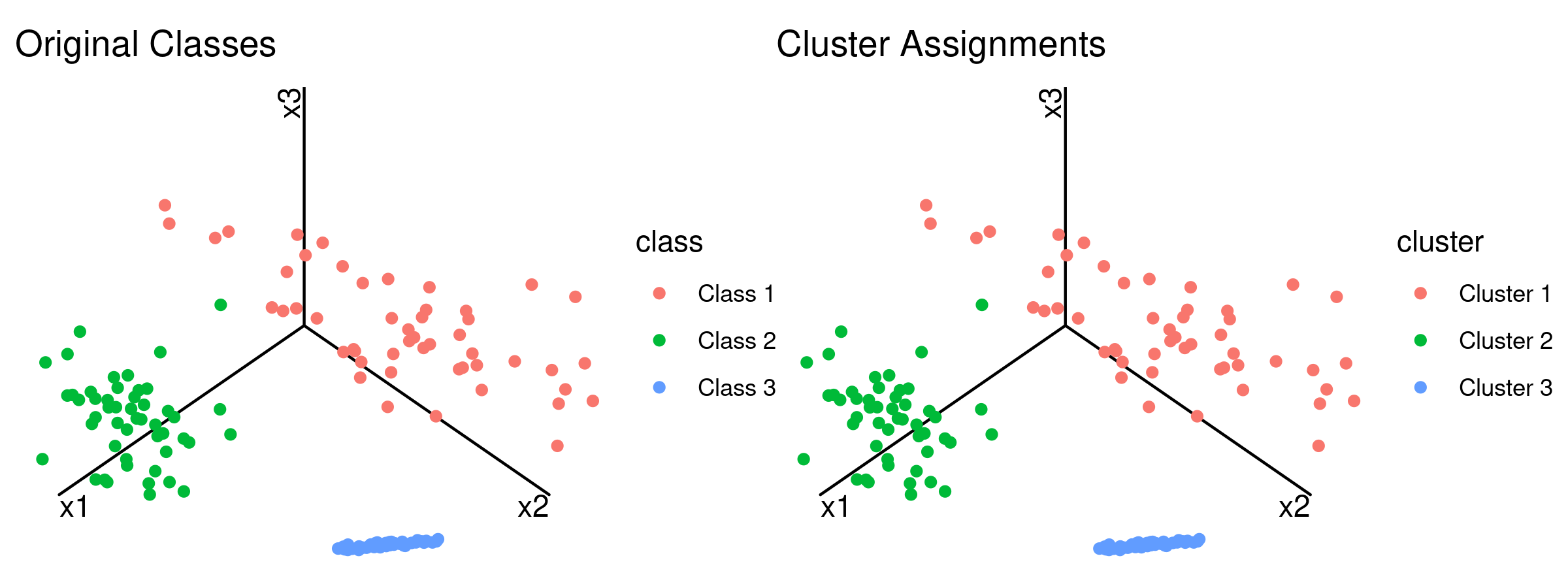}
    \caption{Sampled classes (left plot) versus clusters (right plot).}\label{fig:pdpsim}
\end{figure}

\begin{figure}[h]
    \centering
    \includegraphics[width=0.5\textwidth]{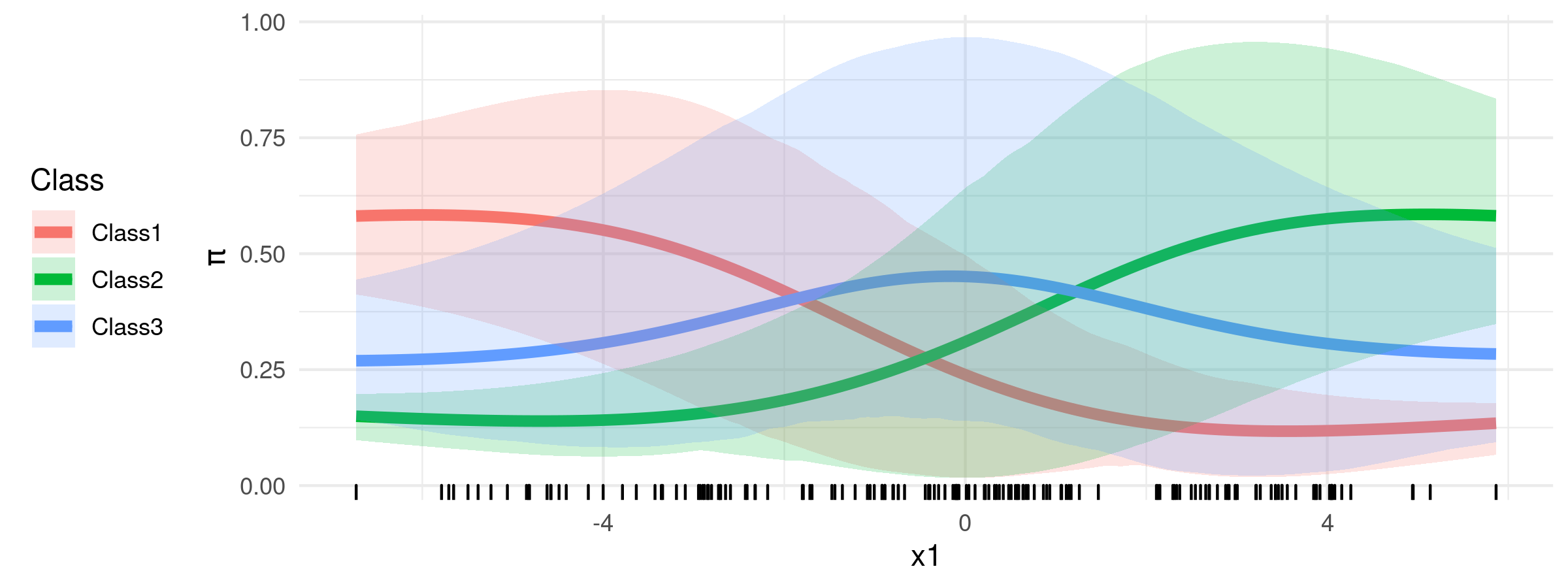}
    \caption{Cluster-specific sPDC curves. Each sPDC curve indicates the average pseudo probability of an observation being assigned to the $c$-th cluster if its $x_1$ value is replaced by the axis value. The bandwidths represent the distribution of sICEC curves the were vertically averaged to the respective sPDC curve.\label{fig:1dpdps}}
\end{figure}

The curves in Fig.  \ref{fig:1dpdps} represent the cluster-wise sPDC. The bandwidths represent 60 percent of the sICE curve ranges that were averaged to receive the respective sPDC. We can see that - on average - $x_1$ has a substantial effect on the clustering outcome. The lower the value of $x_1$ that is plugged into an observation, the more likely it is assigned to cluster 1, while for larger values of $x_1$ it is more likely to be assigned to cluster 2. For $x_1 \approx 0$, observations are more likely to be assigned to cluster 3. The large bandwidths indicate that the clusters are spread out and plugging in different values of $x_1$ into an observation has widely different effects across the data set.
Particularly around $x_1 \approx 0$, where cluster 3 dominates, the average effect loses its meaning due to the underlying sICEC curves being highly heterogeneous. In this case, one should be vary of the interpretative value of the PDC.

\begin{figure}[h]
    \centering
    \includegraphics[width=0.5\textwidth]{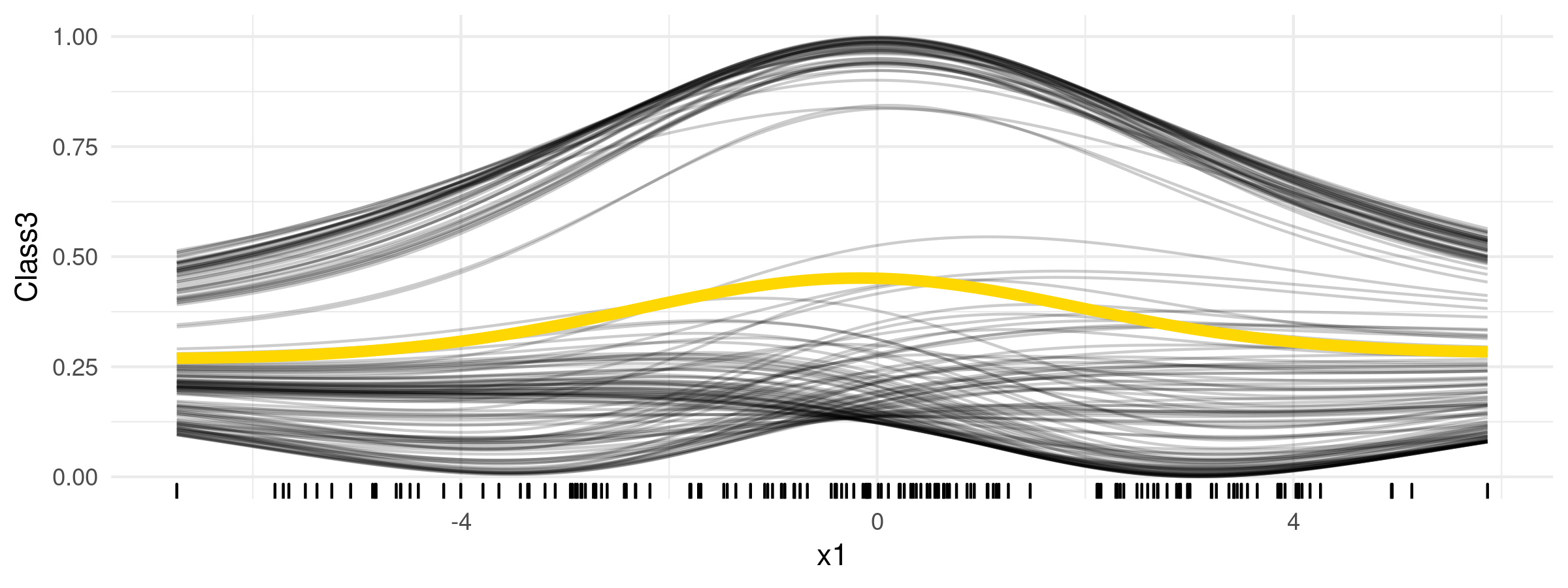}
    \caption{Cluster-specific PDC and ICEC curves, indicating effects on the pseudo probabilities for observations to be assigned to cluster 3.\label{fig:pdcice}}
\end{figure}

We proceed to investigate the heterogeneity of the sICEC curves for cluster 3 (see Fig.  \ref{fig:1dpdps}). Note that the yellow line in Fig. \ref{fig:pdcice} is the blue line in Fig.  \ref{fig:1dpdps}, and the black lines correspond to the sICEC curves that form the blue ribbon in Fig. \ref{fig:1dpdps}.
 The flat shape of the cluster-specific sPDC indicates that $x_1$ has a rather low effect on observations being assigned to cluster 3. However, the sICEC curves reveal that individual effects cancel each outher out when being averaged.

\begin{figure}[h]
    \centering
    \includegraphics[width=0.5\textwidth]{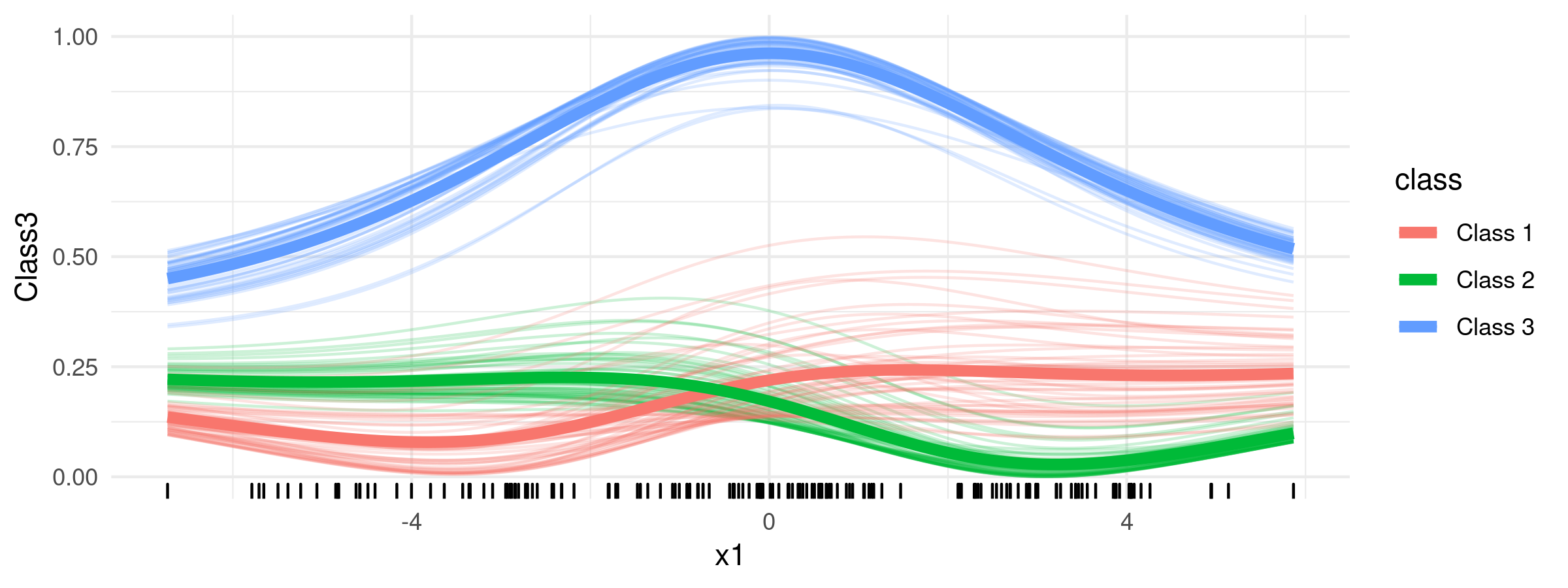}
    \caption{sICEC curves colored by initial cluster assignment. We can see similar effects of replacing the values of $x_1$ on the pseudo probabilities, depending on what initial cluster an observation is part of.\label{fig:cpdcice}}
\end{figure}

It seems likely that observations belonging to a single cluster in the initial clustering run would behave similarly once their feature values were changed. We color each sICEC curve by the original cluster assignment (see Fig. \ref{fig:cpdcice}) and add additional sPDC curves for each initial cluster assignment.
Our assumption - that observations within a cluster behave similarly once we make isolated changes to their feature values - is confirmed.

\subsection{Real Data}

\label{sec:wisconsin_application}

\begin{figure}[h]
    \centering
    \includegraphics[width=0.5\textwidth]{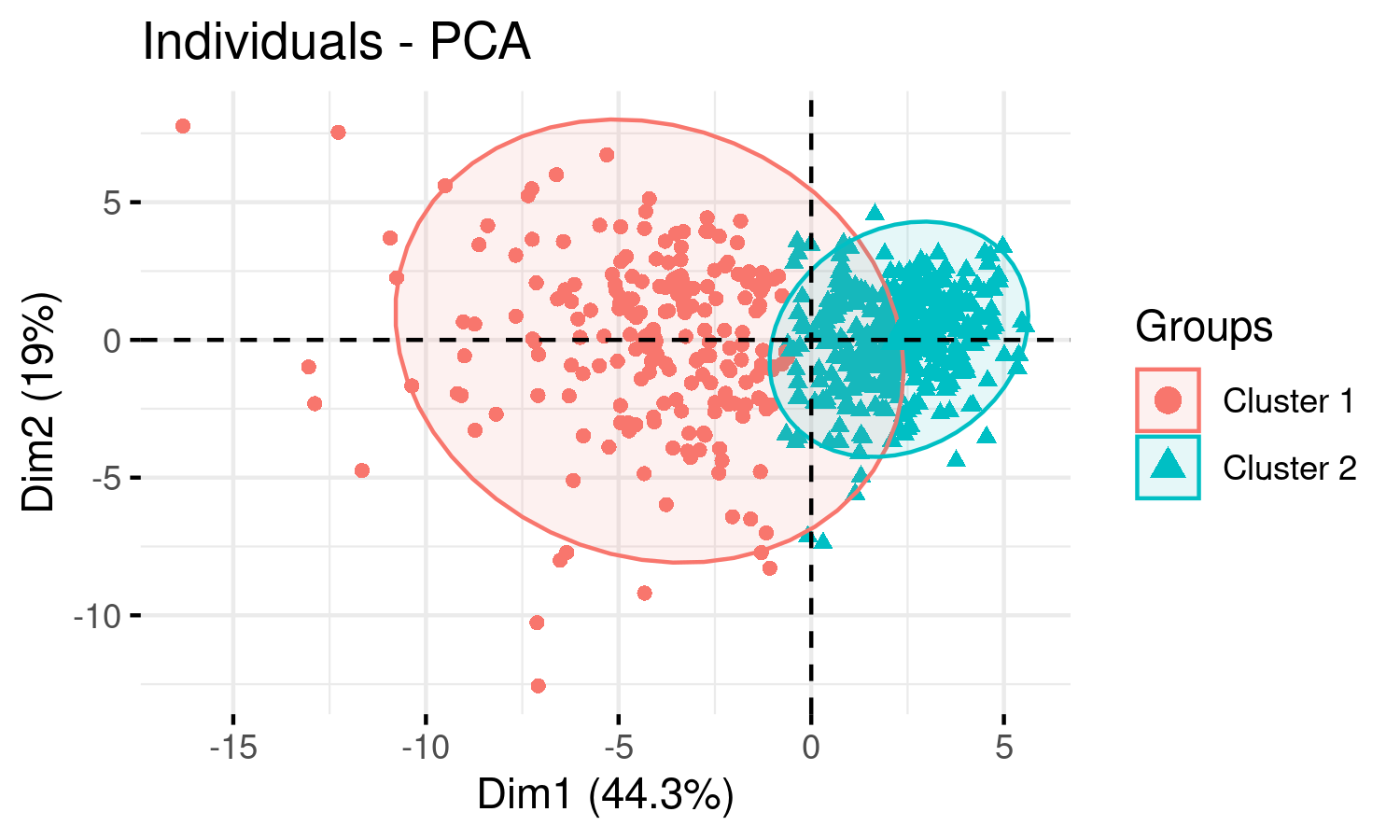}
    \caption{First and second principal components of Wisconsin breast cancer data with clusters of real target values.\label{fig:pca}}
\end{figure}

The Wisconsin diagnostic breast cancer data set \cite{dua_uci} consists of 569 instances of cell nuclei obtained from breast mass. Each instance consists of 10 characteristics derived from a digitized image of a fine-needle aspirate. For each characteristic, the mean, standard error and \enquote{worst} or largest value (mean of the three
largest values) is recorded, resulting in 30 features of the data set.
Each nucleus is classified as malignant (cancer, class 1) or benign (class 2). We cluster the data using Euclidean optimized c-means.
Fig.  \ref{fig:pca} visualizes the projection of the data onto the first two PCs. The clusters cannot be separated with two PCs, and the visualization is of little help in understanding the influence of the original features on the clustering outcome. 

\subsubsection{PFIC}

\begin{figure}[h]
    \centering
    \includegraphics[width=0.5\textwidth]{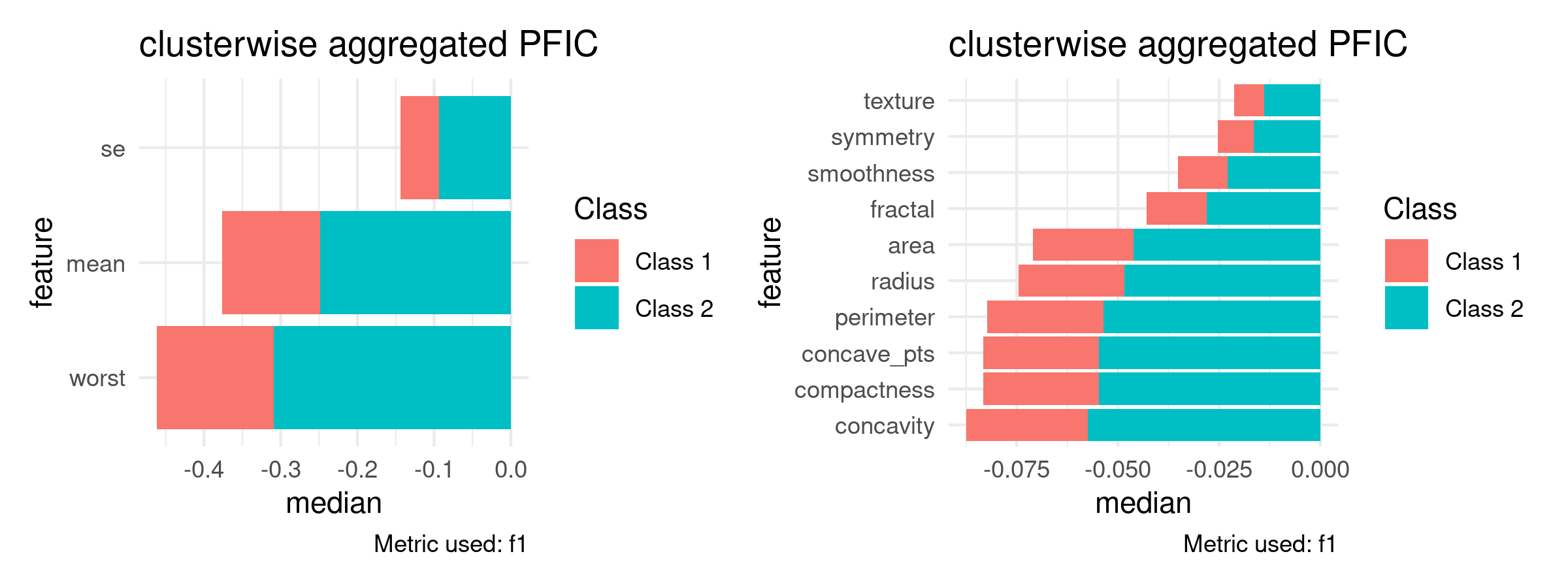}
    \caption{Grouped PFIC (using the natural logarithm) per cluster for groups of categories (left plot) and groups of characteristics (right plot).\label{fig:gFI}}
\end{figure}

We first showcase how the PFIC can serve as an approximation of the actual reclustering, thus accurately representing the influence of features on the clustering outcome (see Table \ref{tab:scoreWDBCI} below and Fig. \ref{fig:wisconsin_pfic} in the Appendix). We use the median F1 score. The first row indicates the performance of the initial clustering run (measured on the latent target variable). Then we recluster the data, once with the 4 most important (second row) and once with the 4 least important features (third row). Dropping the 26 least important features only reduces accuracy by 3\% (measured using the hidden target). In contrast, using the 4 least important features reduces accuracy by 40\%, thus altering the clustering in a major way. This demonstrates that assigning new instances to existing clusters can serve as an efficient method for feature selection. To showcase the grouped feature importance, we jointly shuffle features and compare their importance in Fig. \ref{fig:gFI}. Note that we use the natural logarithm of the PFIC in Fig. \ref{fig:gFI} for better visual separability and to receive a natural ordering of the feature importance (due to F1 being a similarity index), where lower values indicate a lower importance and vice versa.

\begin{table}[h] 
\centering
 \caption{C-means clustering and performance measured on latent target variable for different sets of features.\\MCC = Matthews correlation coefficient}
    \label{tab:scoreWDBCI}
\begin{tabular}{p{1cm}|p{2cm}p{2cm}p{2cm}}
  \hline
 & accuracy & F1 score & MCC\\ 
  \hline
  (1) &0.92 & 0.88 & 0.82\\
  (2) & 0.89 & 0.85 & 0.76\\
  (3) & 0.52 & 0.33 & -0.05 \\
   \hline
\end{tabular}
\end{table}

\subsubsection{ICEC and PDC}

\begin{figure}[h]
    \centering
    \includegraphics[width=0.5\textwidth]{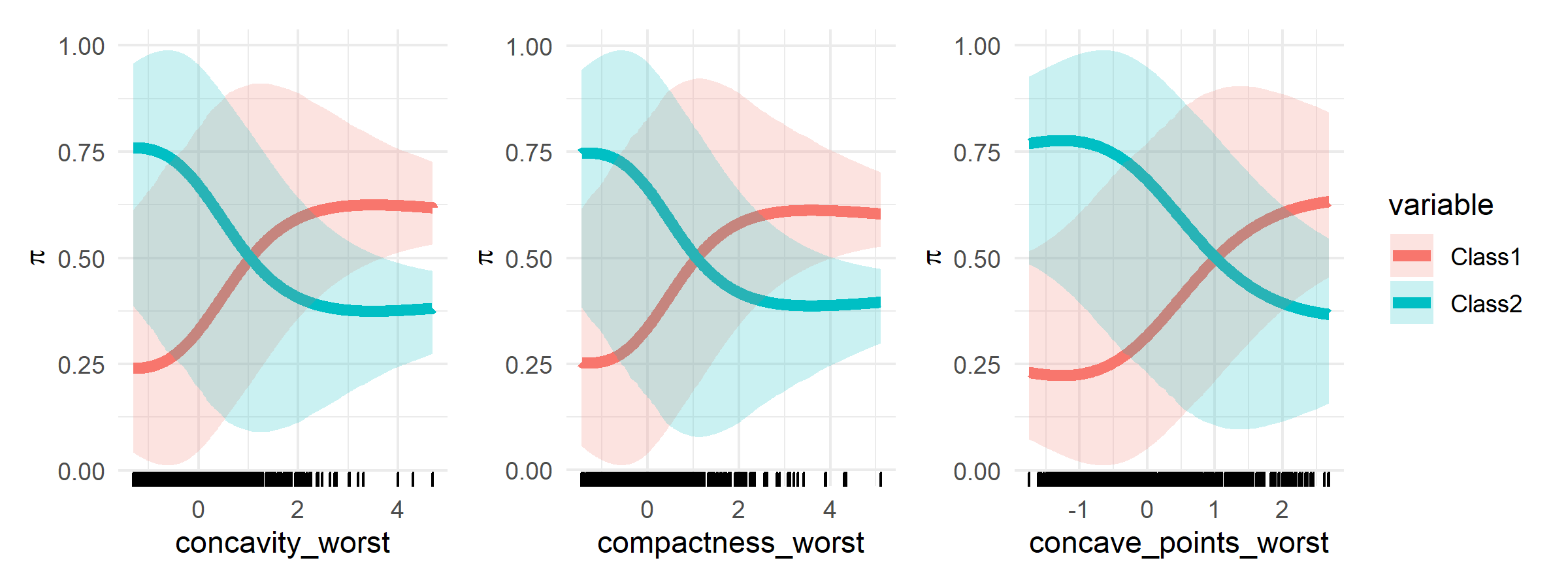}
    \caption{Univariate sPDC plots for the features \texttt{concavity\_worst}, \texttt{compactness\_worst} and \texttt{concave\_points\_worst}.}\label{fig:pdWDBS}
\end{figure}

Fig.  \ref{fig:pdWDBS} plots the sPDC for 3 features \texttt{concavity\_worst}, \texttt{compactness\_worst} and \texttt{concave\_points\_worst}.
The transparent areas indicate the regions where 70\% of the sICEC mass is located. A rug on the horizontal axis shows the distribution of the corresponding feature.
For all three features, larger values result in the observation being assigned to cluster 1, while lower values result in the observation being assigned to cluster 2. The distribution of cluster-specific sICEC curves is large, reflecting voluminous clusters. All features have a strong univariate effect on the clustering, which indicates a large importance. In other words, changing any of these three features would result in a different clustering outcome.

\begin{figure}[h]
    \centering
    \includegraphics[width=0.5\textwidth]{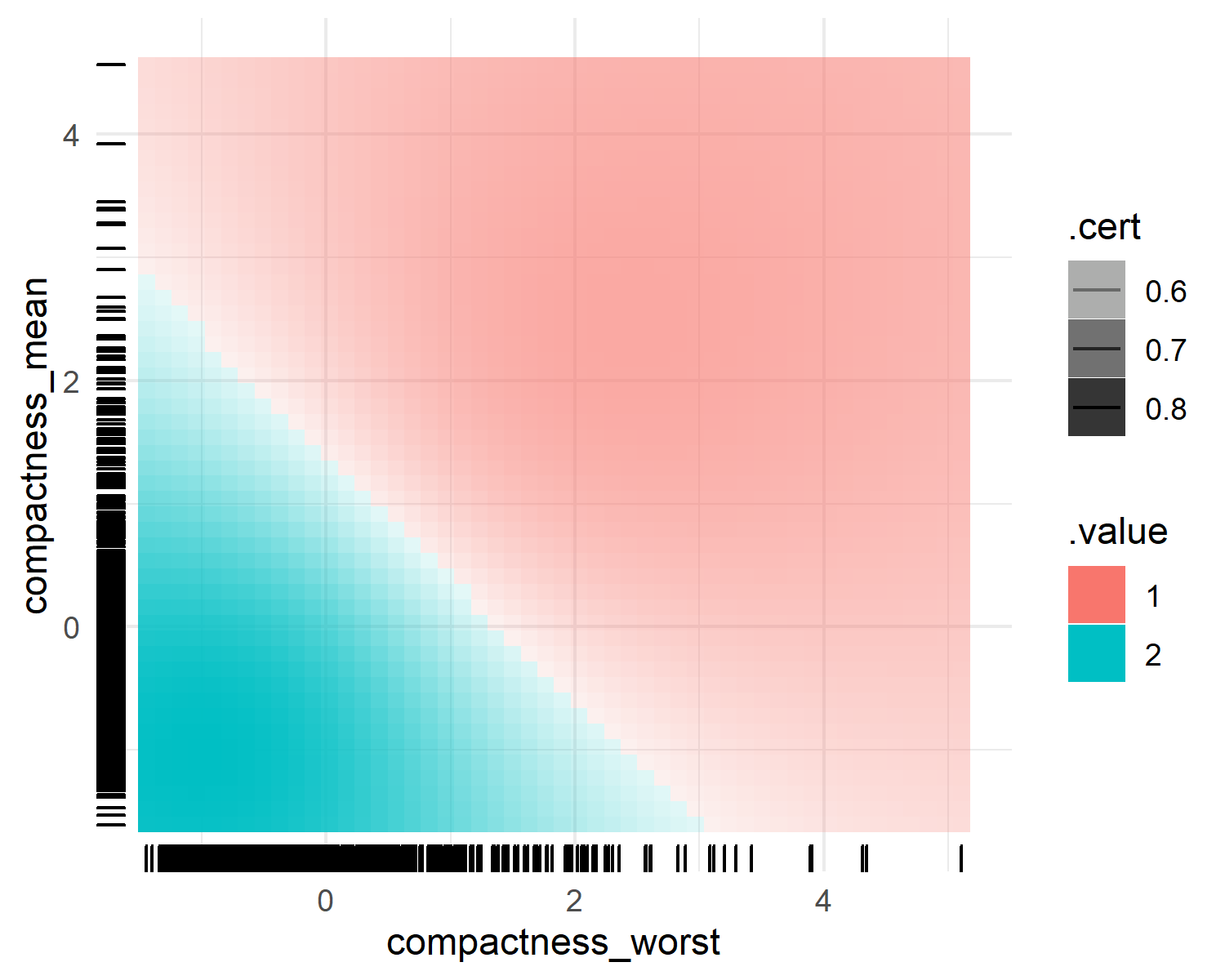}
    \caption{Two-dimensional sPDC for the features \texttt{compactness\_worst} and \texttt{compactness\_mean}. On average, an observation is assigned to cluster 1 for large values of both features, while it is assigned to cluster 2 for low values of both features. This reveals a slight interaction between both features on the clustering outcome.} \label{fig:pd2WDBS}
\end{figure}

Fig.  \ref{fig:pd2WDBS} plots the two-dimensional sPDC for \texttt{compactness\_worst} and \texttt{compactness\_mean}. The color indicates what cluster the observations are assigned to on average when \texttt{compactness\_worst} and \texttt{compactness\_mean} are replaced by the axis values. The transparency indicates the size of the pseudo probability, i.e., the \enquote{certainty} in our estimate. On average, the observations are assigned to cluster 2 when adjusting both features to lower values and to cluster 1 when adjusting both features to higher values. Only a few observations (outliers) are actually located in the vicinity of cluster 1, which indicates cancer. Our methods naturally reveal the nature of the data, i.e., that the Wisconsin diagnostic breast cancer data set represents an anomaly detection problem (detecting cancerous tissue). 

\section{Conclusion}
\label{sec:conclusion}

This research paper presents three interpretation methods suited for any clustering algorithm able to reassign instances through soft or hard labels: The PFIC, ICEC, and PDC. Our methods characterize the relevance of features for the entire clustering outcome or the constitution of clusters in lower dimensions. The PFIC is a general framework that outputs a single, global value for each feature indicating its importance to the clustering outcome or one value for each cluster (and feature). It can be easily adjusted to the data and algorithm at hand due to its flexibility in terms of different score metrics and the possibility for global and cluster-specific assessments. The ICEC and PDC add to these capabilities by visualizing the structure of the feature influence on the clustering across the feature space for single observations and the entire feature space. We argue that interpretations based on SA of the data are superior to actual dimension reduction, as the clustering is explained in the same space as it was created. However, they do not replace interpretable clustering algorithms (e.g., which actively search for interpretable clusters and/or explain clustering decisions and what distinguishes clusters from each other) but rather complement them. First, the PFIC, ICEC, and PDC can also assist in conducting interpretations for interpretable algorithms. Second, there are few alternatives to our methods for explaining an outcome of a non-interpretable clustering algorithm.
\par
Although explaining algorithmic decisions is an active research topic in SL, it is largely ignored in unsupervised learning, including clustering. Our proposed methods add to the limited works on cluster interpretability, specifically on algorithm-agnostic interpretation methods. With this research paper, we hope to demonstrate the untapped potential of adapting existing techniques in SL to the unsupervised setting and spark more research in this direction.

\section*{Credit Taxonomy}

Conceptualization: Henri Funk (HF), Christian Scholbeck (CS), Giuseppe Casalicchio (GC); Methodology: HF, CS, GC; Software and Validation: HF; Formal Analysis and Investigation: HF, CS, GC; Writing - Original Draft: HF, CS; Writing - Review and Editing: HF, CS, GC; Visualization: HF; Supervision: CS, GC; Project Administration: GC

\newpage
\bibliographystyle{IEEEtran}
\bibliography{bibfile} 

\newpage
\appendix

\subsection{Binary Scores}\label{sec:bsf}

\begin{itemize}
\item 
$F_\beta$ score: Balances false positives and false negatives.

\vspace{0.25cm}

Relating to Section \ref{sec:pfi}, the $F_\beta$ score of cluster $c$ versus the remaining ones
corresponds to:
$$
F_{\beta, c} =\frac{\left(\beta^{2}+1\right) \cdot P_c \cdot R_c}{\beta^{2} \cdot P_c+R_c}\\
$$
where
\begin{align*}
    P_c &=\frac{\#_{cc}}{\#_{cc}+\#_{\overline{c}c }} \\
    R_c &=\frac{\#_{cc}}{\#_{cc}+\#_{c\overline{c}}}
\end{align*}
The $F_1$ (which we refer to as F1) score simplifies to:
$$
F_{1, c} = 2 \frac{P_c \cdot R_c}{P_c + R_c}\\
$$
\item Jaccard Index:
Identifies equivalency between data sets.
\begin{equation*}\label{eq:ji}
    J_c = \frac{\#_{cc}}{\#_{cc}+\#_{\overline{c}c }+\#_{c\overline{c}}}
\end{equation*}
\item
Folkes-Mallows index:
Identifies similarities between clusters.
\begin{equation*}\label{eq:fmi}
FM_c=\sqrt{\frac{\#_{cc}}{\#_{cc}+\#_{\overline{c}c }} \cdot \frac{\#_{cc}}{\#_{cc}+\#_{c\overline{c}}}}
\end{equation*}
 
\end{itemize}

\subsection{Multi-Class Scores}

Let $s_\text{bin, i}$ be an arbitrary binary score for the $i$-th class and $w_i$ be the proportion of class $i$ in the data set. $s_\text{macro}$ denotes the multi-class macro score that treats each cluster with equal importance. $s_\text{micro}$ denotes the multi-class micro score that treats each instance with equal importance:
\begin{align*}
    s_\text{macro} &= \frac{1}{k} \sum_{i = 1}^k s_\text{bin, i} \\
    s_\text{micro} &= \sum_{i = 1}^k w_i \;  s_\text{bin, i}
\end{align*}

\subsection{Wishart Distribution}\label{sec:wish}

We sample a covariance matrix $M$ from the Wishart distribution
$M \sim \text{Wishart}_{3}(3, \Sigma)$, where $\Sigma$ is constructed as follows:

\begin{align*}
        \Sigma_\text{Class 1} &= 0.6 I_3 \\
        \Sigma_\text{Class 2} &= 0.3 I_3 \\
        \Sigma_\text{Class 3} &= 0.15 I_3
\end{align*}
$I_3$ refers to the $3 \times 3$ identity matrix.
As a result, the variance of class 1 is the largest, the variance of class 3 is the lowest, and the variance of class 2 lies between the variance of class 1 and 3. This results in the following distributions:
\begin{equation*}\label{eq:pdpsim}
{\scriptsize
    \begin{array}{rcl}
    \left( \begin{array}{c} x_1 \\ x_2 \\x_3 \end{array} \right)_\text{class 1} & \sim  
    \mathcal{N}\left(\left(\begin{array}{c} -3 \\ 0\\ 3\end{array}\right),
    \left(\begin{array}{ccc}
    2.23 & -2.30 & 0.49 \\ 
    -2.30 & 5.07 & -0.83 \\ 
    0.49 & -0.83 & 1.65 \\  \end{array}\right)\right)\\[1.5em]
    \left( \begin{array}{c} x_1 \\ x_2 \\x_3\end{array} \right)_\text{class 2} & \sim  
    \mathcal{N}\left(\left(\begin{array}{c} 3 \\ -3\\ 0 \end{array}\right),
    \left(\begin{array}{ccc}
    1.22 & -0.21 & -0.21 \\ 
    -0.21 & 0.77 & -0.58 \\ 
    -0.21 & -0.58 & 0.55 \\ \end{array}\right)\right)\\[1.5em]
    \left( \begin{array}{c} x_1 \\ x_2 \\ x_3 \end{array} \right)_\text{class 3} & \sim  
    \mathcal{N}\left(\left(\begin{array}{c} 0 \\ 3\\ -3 \end{array}\right),
    \left(\begin{array}{ccc} 
    0.60 & -0.25 & 0.39 \\ 
    -0.25 & 0.20 & 0.14 \\ 
    0.39 & 0.14 & 1.20 \\\end{array}\right)\right)
    \end{array}
}
\end{equation*}

\vspace{0.5cm}
\subsection{PFIC for Wisconsin Diagnostic Breast Cancer Data}

\begin{figure}[h]
    \centering
    \includegraphics[width = 0.5\textwidth]{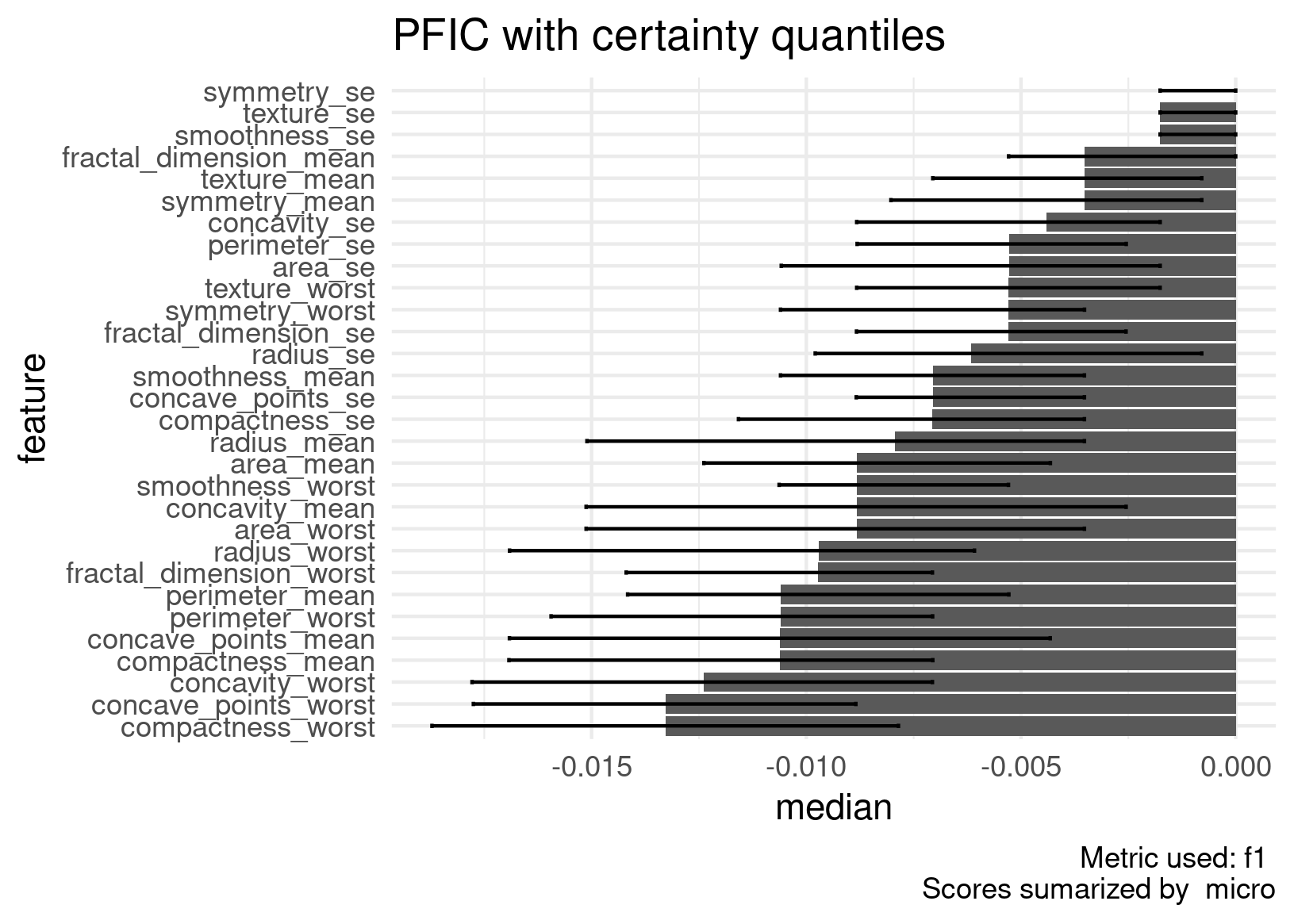}
    \includegraphics[width = 0.5\textwidth]{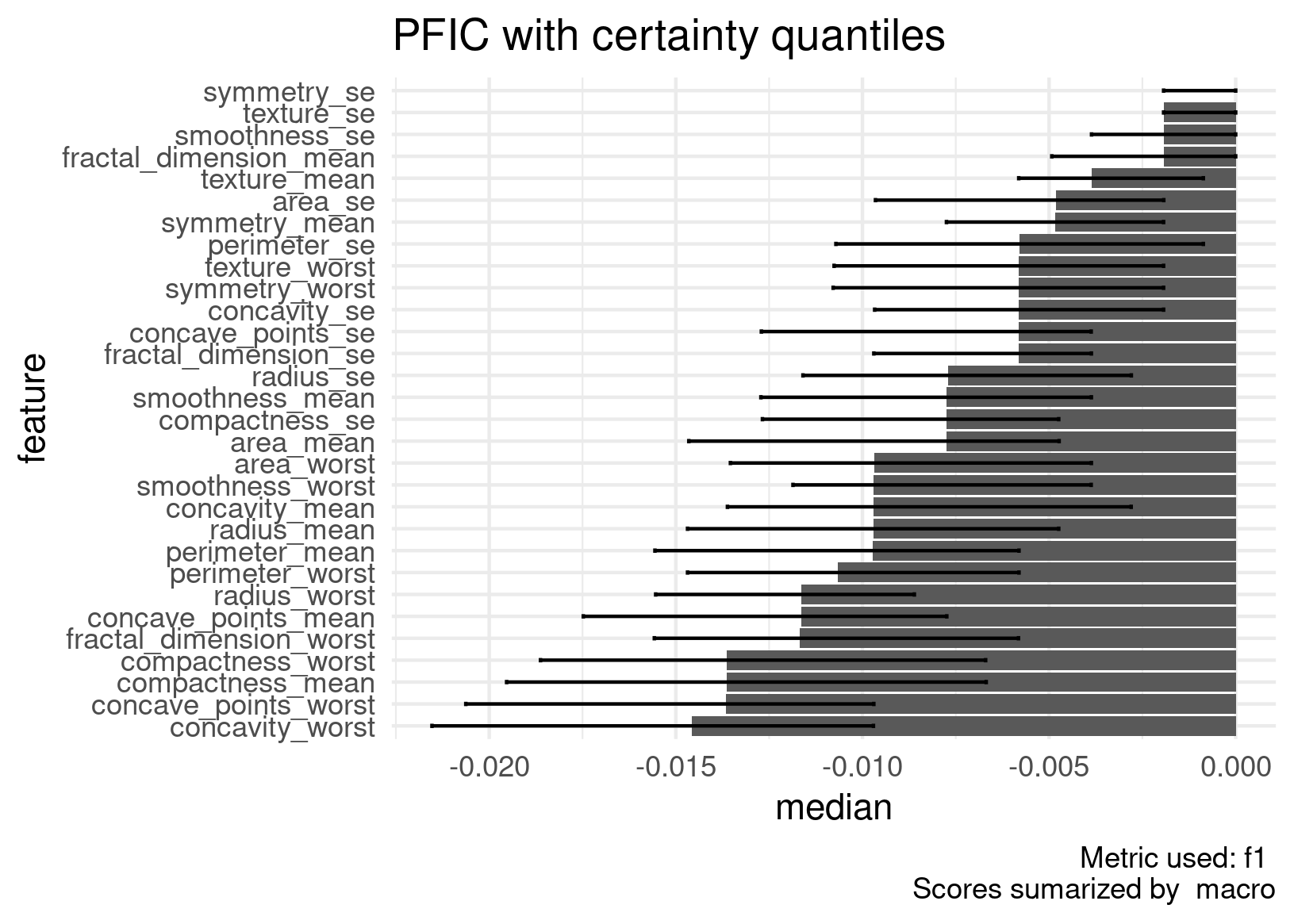}
    \caption{Global PFIC on Wisconsin diagnostic breast cancer data clustering using micro (top) and macro (bottom) F1 score. We use the natural logarithm of the PFIC for better visual separability and to receive a natural ordering of the feature importance, where lower values indicate a lower importance and vice versa. This reverses the ordering of the F1 score which is a similarity index, where lower values indicate higher dissimilarity between shuffled and unshuffled data and thus a high importance. \label{fig:wisconsin_pfic}}
\end{figure}

\end{document}